\newif\ifanon
\anonfalse

\def\paperTitle{The Open Motion Planning Library 2.0}

\ifanon
    \def\paperAuthor{Anonymous Author(s)}
\else
    \def\paperAuthor{
    Weihang Guo$^1$,
    \and
    Theodoros Tyrovouzis$^1$,
    \and 
    Emiliano Flores$^1$,
    \and 
    Clayton W. Ramsey$^1$, 
    \and
    Zachary K.\ Kingston$^{2*}$,
    \and
    Ioan~A.~\c{S}ucan$^{3*}$,
    \and
    Mark Moll$^4$,
    \and
    Lydia E.\ Kavraki$^{1,5}$
    \thanks{$^1$WG, TT, EF, CWR, LEK are with the Department of Computer Science, Rice University, Houston TX, USA. {\tt\footnotesize \{wg25,tt88,ef55,cwr3, kavraki\}@rice.edu}}
    \thanks{$^2$ZKK is with the Department of Computer Science, Purdue University, West Lafayette, IN. {\tt\footnotesize zkingston@purdue.edu}} 
    \thanks{$^3$IA\c{S} is with Waymo, LLC, Mountain View, CA}
    \thanks{$^4$MM is with Metron, Inc., Reston, VA}
    \thanks{$^5$LEK is also affiliated with the Ken Kennedy Institute at Rice University}
    \thanks{$^*$Work by these authors was done while they were at Rice University.}
    }
\fi
\documentclass[letterpaper]{IEEEtran}
\usepackage{amsmath,amssymb,amsfonts}
\usepackage{graphicx}
\usepackage{textcomp}
\usepackage{xcolor}
\usepackage{amsmath}
\usepackage{amsthm}
\usepackage{algorithm}
\usepackage{subfig}
\usepackage{algpseudocode}
\usepackage{grffile}
\usepackage{wrapfig,epsfig}
\usepackage{url}
\usepackage{xcolor}
\usepackage{epstopdf}
\usepackage{bbm}
\usepackage{dsfont}
\usepackage{booktabs}
\usepackage{listings}
\usepackage{tikz}
\usetikzlibrary{positioning}

\usepackage[
    activate   = {true},
    protrusion = false,
    expansion  = true,
    kerning    = true,
    spacing    = true,
    tracking   = false,
    auto       = true,
    selected   = true,
    factor     = 1000,
    stretch    = 50,
    shrink     = 10,
]{microtype}
\usepackage[pdfa,colorlinks,bookmarksopen,bookmarksnumbered,allcolors=darkgray]{hyperref}
\usepackage[
    maxbibnames=99,
    maxcitenames=2,
    natbib=true,
    style=numeric-comp,
    backend=biber,
    sorting=none,
    giveninits=true,
    url=false, 
    doi=false,
    eprint=false,
    isbn=false,
]{biblatex}

\graphicspath{{./}}

\usepackage{balance}
\usepackage{lastpage}
\makeatletter
\def\lastreferencepage{\lastpage@lastpage} %
\makeatother
\def\balanceissued{unbalanced}%
\renewbibmacro{finentry}{%
    \ifnum\thepage=\lastreferencepage%
        \expandafter\ifx\expandafter\relax\balanceissued\relax%
        \else%
            \balance%
            \gdef\balanceissued{\relax}%
        \fi%
    \fi%
    \finentry%
}

\begin{document}

\title{\paperTitle}
\author{\paperAuthor}
\maketitle
\begin{abstract}
The Open Motion Planning Library (\textsc{OMPL}), first released in 2008, has become a cornerstone of the motion planning community, providing implementations of a wide range of state-of-the-art sampling-based algorithms. Over almost two decades of continuous development, we have steadily expanded the library with new planners, state spaces, and problem formulations. These additions range from asymptotically optimal and lazy planners to constrained motion planning and planning with temporal-logic goals. Building on this foundation, we introduce OMPL~2.0, a major evolution of the library that targets real-time motion planning through hardware acceleration and integrates seamlessly with modern AI research workflows. We also reflect on how OMPL and the field of motion planning have grown together over the years, and discuss the library's broader impact on the research community.

\end{abstract}

\section{Introduction}

The history of motion planning can be traced back to the late 1970s, when the concept of configuration space~\cite{Lozano1983cspace} was introduced to formally describe robot motion planning problems. Early approaches focused on exact algorithms~\cite{canny1988complexity} and potential field methods~\cite{barraquand1991numerical, bhatia2011motion}. However, these techniques often struggled with the high computational complexity of realistic robotic systems, as research has proven that motion planning in configuration space is NP-hard~\cite{canny1988complexity}. In the 1990s, sampling-based motion planning emerged as a practical approach. Algorithms such as the probabilistic roadmap~(PRM)~\cite{kavraki1996prm}, expansive-space tree~(EST)~\cite{hsu1999path}, and rapidly-exploring random trees~(RRT)~\cite{kuffner2000rrt} developed before 2000, approximate the connectivity of the configuration space through random sampling and enable efficient planning in high-dimensional spaces. These approaches quickly became a dominant paradigm in robotics. Several early sampling-based planners were accompanied by theoretical analyses that characterized their trade-offs~\cite{kavraki1998analysis, kavraki1995randomized}. Subsequent research introduced improved sampling strategies and algorithmic refinements. Later, in the 2010s, asymptotically optimal planners such as PRM*, RRT*~\cite{karaman2011sampling}, and SST~\cite{li2016asymptotically-optimal-sampling-based-kinodynamic} provided convergence guarantees to optimal solutions.
As sampling-based motion planning algorithms rapidly developed, implementing and fairly comparing new planners became increasingly challenging. Researchers often needed to reimplement existing algorithms and supporting data structures, making reproducibility and benchmarking difficult. 

To address the reproducibility and benchmarking challenges, the Open Motion Planning Library~(OMPL)~\cite{sucan2012the-open-motion-planning-library} was developed and released in 2008 as an open-source software framework for motion planning. OMPL provides implementations of a wide range of sampling-based planners together with reusable and flexible abstractions for defining planning problems, enabling researchers and practitioners to easily evaluate algorithms, benchmark with other planners, and integrate motion planning into robotics applications.

Almost two decades have passed since OMPL's first release. Over this period, motion planning research, computer architectures, and the broader robotics ecosystem have all advanced substantially. OMPL has evolved alongside these changes, continually incorporating new planners, state space representations, and problem formulations. In this paper, we introduce OMPL~2.0, a major evolution of the library that targets real-time motion planning through hardware acceleration and integrates with modern AI research workflows.

\section{The Open Motion Planning Library}

Before introducing new features of OMPL~2.0, we recap the main components of OMPL. This design, introduced in the original OMPL paper~\cite{sucan2012the-open-motion-planning-library}, remains largely unchanged and has stood the test of time. OMPL is structured as a set of modular components that correspond closely to the fundamental concepts of sampling-based motion planning. The core abstractions include state space, state validator, state sampler, start and goal, and motion planners, which together define the planning problem.

The \emph{state space} defines the representation and topology of the robot's configuration space, along with its limits and any kinematic or dynamic constraints. A \emph{state validator} provides a generic interface for collision checking, allowing users to connect any collision checking library or simulator. Finally, a \emph{state sampler} draws configurations uniformly or from a Gaussian distribution over the state space. With these foundational components in place, users can select or build their own \emph{motion planner} and specify \emph{start} and \emph{goal} states to solve the problem.
Beyond solving individual planning problems, the \emph{benchmark} component provides extensible infrastructure for running evaluations and analysis across multiple planners~\cite{moll2015benchmarking-motion-planning-algorithms}. Results and metadata are stored in a database to support reproducibility.

A key design choice in OMPL is to focus exclusively on motion planning algorithms and to not include specific representations of robots, workspaces, or collision detection. Instead, OMPL relies on external software components to provide these capabilities. This minimalist design allows the library to remain general and easily integrated with a wide range of robotics systems and simulation environments.
OMPL is a community effort, with contributions from many researchers over the years.

\begin{figure}
    \centering%
    \includegraphics[width=.45\textwidth]{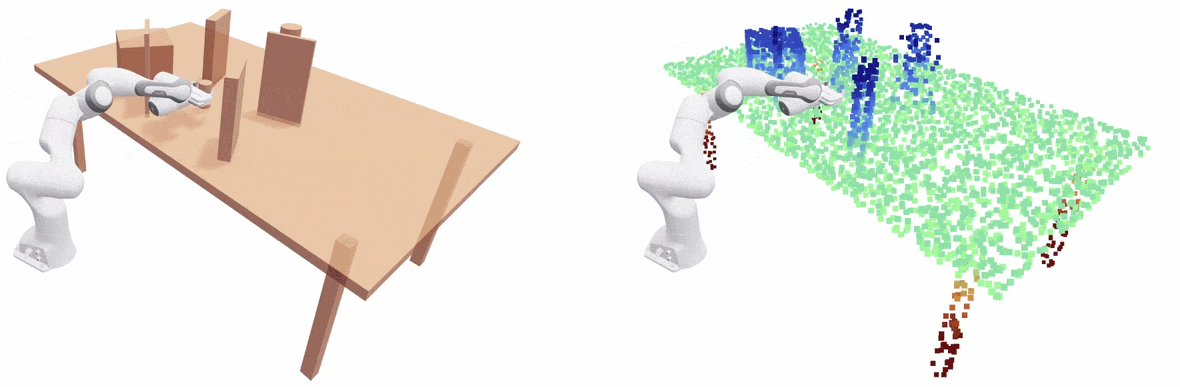}
    \caption{OMPL~2.0 supports planning with realistic robots and environments represented by meshes or point clouds.}\label{fig:ompl2-vamp}
\end{figure}

\section{Two Decades of Adoption}

OMPL has become a widely used tool in the motion planning community. The original paper has accumulated over 2,400 citations, reflecting sustained and widespread adoption. By providing carefully maintained reference implementations of canonical planners, it has helped standardize how the community benchmarks and communicates algorithmic contributions, allowing researchers to focus on novel contributions rather than reimplementing baselines.

A key factor behind this adoption is OMPL's separation from robot models, workspace representations, and collision geometry. By defining a clean interface around state spaces, validity checkers, and planners, OMPL integrates naturally with external systems without imposing constraints on how they are built. OMPL can be called from MoveIt~\cite{chitta2012moveit}, robot simulators such as CoppeliaSim~\cite{rohmer2013coppeliasim} and Flightmare~\cite{song2020flightmare}, industrial frameworks like Tesseract~\cite{tesseract_robotics}, and research toolkits including AIKIDO~\cite{aikido}, EXOTica~\cite{exotica}, 
and the Kautham Project~\cite{rosell2014kautham}. OMPL's planners have also been used in task-and-motion planning~\cite{dantam2018incremental} and mobile robot navigation via Nav2~\cite{macenski2020marathon}, while tools such as HyperPlan~\cite{moll2021hyperplan}, Robowflex~\cite{kingston2021robowflex}, and MotionBenchMaker~\cite{chamzas2021motionbenchmaker} build on its benchmarking infrastructure for standardized algorithm evaluation. OMPL has also been used in non-robotics applications, such as modeling the conformational flexibility of large macromolecules~\cite{gipson-moll2013sims-hybrid-method}.

\section{OMPL 2.0: Key Technical Improvements}\label{sec:tech_improvements}

Rather than emerging from a single landmark release, OMPL has matured through more than a decade of incremental development. Since the publication of the original paper, over 30 versions have been released along the path from OMPL~0.x through OMPL~1.0 to OMPL~2.0, each contributing new algorithms and capabilities. These updates reflect both advances in motion planning research and the evolving needs of the robotics community. 

\paragraph{Problem Set Expansion} OMPL~2.0 now supports a wider range of planning paradigms, including asymptotically optimal planners~\cite{wilson2025aorrtc, gammell2020bitstar} and lazy planning methods~\cite{bohlin2000lazyprm}. Beyond extending its set of algorithms, OMPL~2.0 has also broadened the kinds of planning problems it can express. Constrained motion planning~\cite{kingston2019constrained} enables any sampling-based planner to operate directly on an implicit constraint manifold, supporting practical problems involving contact constraints and closed kinematic chains. For task-level specifications, LTLPlanner~\cite{bhatia2011motion} is a kinodynamic planner that produces trajectories satisfying a given linear temporal logic formula by searching the cross product of the continuous state space and the discrete space of the formula's accepting traces. In addition, the library has expanded its set of state space representations to cover a broader class of problems, including discrete state spaces, time-augmented spaces for planning under temporal constraints, and non-Euclidean spaces such as Dubins and Reeds–Shepp for car-like vehicles. Together, these additions significantly broaden the types of planning problems that can be addressed within OMPL~2.0. The ever-growing number of planning algorithms, each with its own parameters, can make it challenging to select an appropriate planner for a specific robot or environment. We have shown that hyperparameter tuning can be used as an effective tool for planning algorithm selection and tuning~\cite{moll2021hyperplan}.

\paragraph{Performance Evolution} Beyond advances in planning algorithms, OMPL~2.0 integrates VAMP~\cite{thomason2024vamp}, which leverages CPU SIMD instructions to perform collision checking and forward kinematics with fine-grained, hardware-efficient parallelism. With this integration, planners can find solutions in microseconds, and aggregate solution rates can reach the kilohertz range, all without specialized hardware such as a GPU. OMPL now also includes CAPT~\cite{ramsey2024capt}, which extends the SIMD philosophy to point cloud collision checking. These capabilities open up new opportunities for applications that require extremely fast planning in dynamic environments, including reactive planning and task-and-motion planning. The OMPL~2.0's integration with VAMP and CAPT also makes it straightforward to apply motion planning to realistic robots operating in real-world scenes (see Fig.~\ref{fig:ompl2-vamp}).

\paragraph{Modernized Development Infrastructure} When OMPL was first released, many of the software development tools and practices now common in open-source projects were not yet widely adopted. At the time, OMPL used Py++~\cite{pyplusplus} to generate its Python bindings. The tool is no longer maintained. To address this, OMPL~2.0 has transitioned to nanobind~\cite{nanobind}, a modern C++ binding framework that provides a lightweight and efficient interface between C++ and Python while simplifying the maintenance of binding code~\cite{guo2026omplpython}. In addition, OMPL~2.0 has adopted contemporary development infrastructure such as GitHub Actions for continuous integration, enabling automated building and testing across multiple platforms. The project also employs automated pipelines to build and publish Python wheels to PyPI, significantly simplifying installation and distribution.

\paragraph{Streamlining} Earlier versions of OMPL included OMPL.app, a graphical front end that demonstrated integration with external libraries for mesh loading, collision checking, and visualization. Over the past decade, the robotics ecosystem has matured, with widely used simulation platforms and collision detection libraries that integrate naturally with OMPL, making a dedicated graphical interface no longer necessary. OMPL.app has therefore been removed in OMPL~2.0, allowing the library to focus on efficient planning infrastructure and tighter integration with external robotics software.

\section{Project Community and Organization}

The goal from the beginning of the OMPL project has always been to create a resource that is useful for research, education, and real-world use. This goal has been achieved, but it is also something that requires ongoing effort to retain relevance. Long-term software maintenance and support remain a challenge in an academic environment. Shortly after the initial release, we organized a few tutorials at robotics conferences to get researchers started with OMPL and help start a user community. Improved documentation and more general awareness of OMPL have reduced the need for this. We provide the link to the most recent OMPL tutorial.\footnote{\url{https://kavrakilab.org/icra-2026-ompl-tutorial/}}

\begin{figure}
\centering\includegraphics[width=\linewidth]{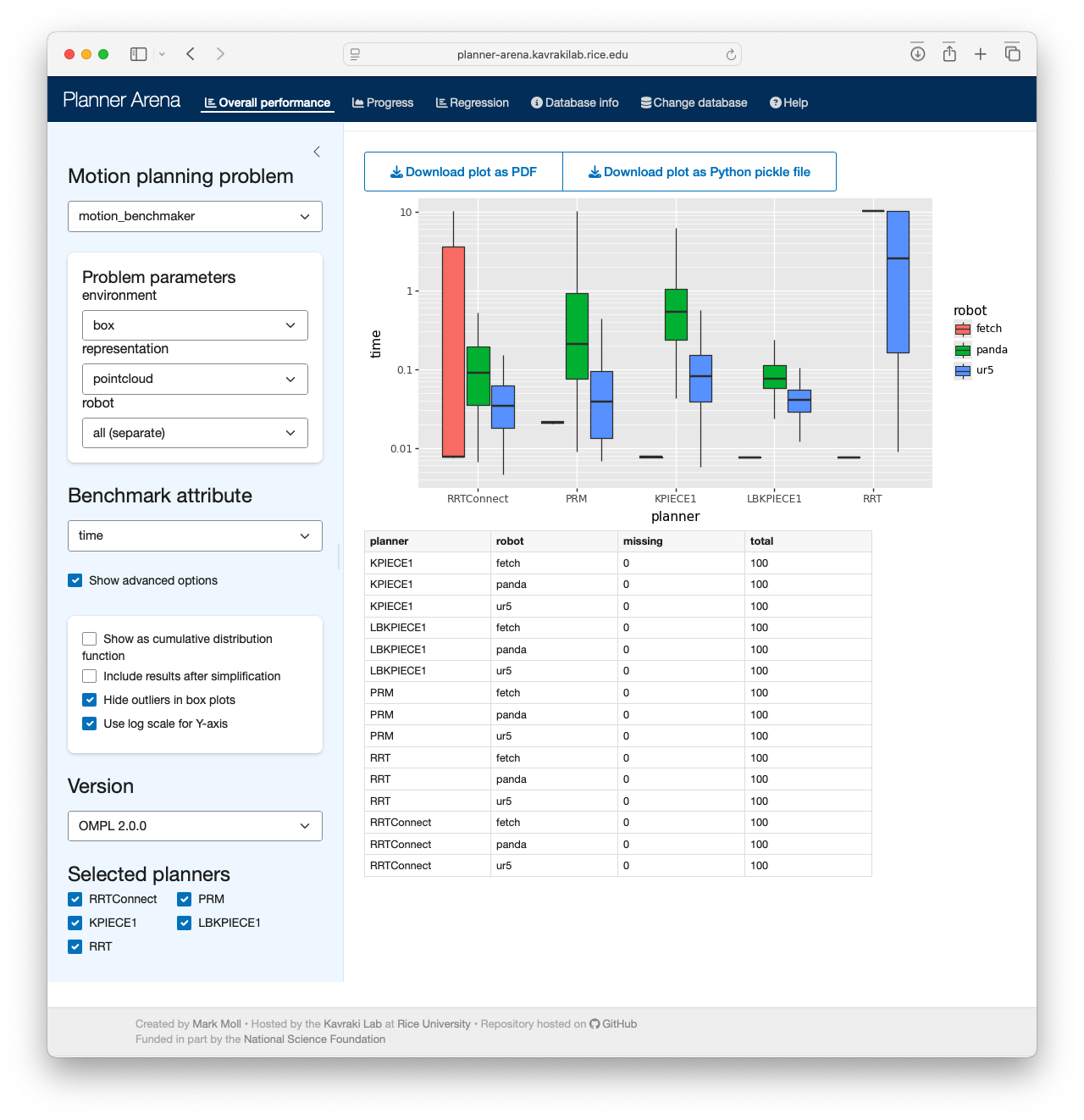}
\caption{Planner Arena allows users for benchmarking motion planning algorithms.}\label{fig:benchmarking}
\end{figure}
We also decided early on to spend significant time on building an extensible infrastructure for benchmarking motion planning algorithms~\cite{moll2015benchmarking-motion-planning-algorithms}. This has been invaluable in quantifying how new planners advance the state of the art. More recently, we have also generated large datasets of realistic hard motion planning problems (and made the tooling for generating such datasets available as well)~\cite{chamzas2021motionbenchmaker}. Benchmark results can be visualized through an interactive browser-based tool called Planner Arena.\footnote{\url{https://planner-arena.kavrakilab.rice.edu/}} Fig.~\ref{fig:benchmarking} shows sample results for a parameterized benchmark. The parameters in this case are (1) the specific motion planning problem, (2) how the environment is represented (mesh vs.~point cloud), and (3) the type of robot (a Fetch, a Panda, and a UR5 in this example). The UI enables comparison of planning algorithms on all robots for a single problem, on a single robot on all problems, etc. In each case, there are many performance metrics that can be visualized, such as planning time and solution path length. The table in the bottom right shows the number of missing values: if a planner was not able to find a path on some runs, the path length would be considered missing for those runs. In summary, Planner Arena encourages exploration and a nuanced characterization of planner performance. 

The abstractions that OMPL imposes on planning algorithms are minimal and make it straightforward to develop prototypes of planners. Evaluation in simplified environments (e.g., a point robot in 2D) makes testing easy, before deployment in realistic environments without having to change the planning algorithm implementation itself.

Many motion planning researchers have developed new motion planning algorithms using OMPL. Oftentimes, the original authors contributed their implementations to the OMPL repository. For example, Karaman and Frazzoli contributed the initial RRT* implementation to OMPL shortly after publication~\cite{karaman2011sampling}. Bekris et al.\ contributed implementations of their asymptotically near-optimal algorithms~\cite{li2016asymptotically-optimal-sampling-based-kinodynamic}. Gammell et al.\ have contributed many algorithms such as BIT*, ABIT*, AIT*, EIT*, and AORRTC~\cite{strub2022ait-eit,wilson2025aorrtc}. Orthey et al.\ have contributed implementations of their work on multi-level planning~\cite{orthey2024multilevel} and planning in space-time~\cite{Grothe2022ICRA}. This is by no means an exclusive list.\footnote{A list of contributors is maintained in \url{https://ompl.kavrakilab.org/developers.html}}

Over time, the OMPL community has grown to include many AI researchers, who often use OMPL simply as a black box. Because Python is the primary language for integrating software components in the AI community, OMPL's Python bindings have become increasingly important. Originally, these bindings were intended mainly to lower the barrier to entry for beginning programmers. Today, their performance penalty is small enough to be acceptable for many use cases, though we still recommend the C++ API directly when optimal performance is needed.

\section{The Next Decade of OMPL}

Looking ahead, we see several directions that will shape the future development of OMPL. Motion planning is increasingly used as a component within larger decision-making systems: task and motion planning~\cite{dantam2018incremental} combines symbolic reasoning with geometric planning, planning under uncertainty, such as POMDPs~\cite{liang2024scaling}, demands rapid replanning as beliefs evolve, and reactive planning~\cite{bhardwaj2022storm} requires algorithms that quickly adapt to changing environments. Supporting these paradigms will require continued investment in fast, anytime planning and interfaces that facilitate integration with higher-level reasoning. Beyond robotics pipelines, OMPL can also serve the broader AI ecosystem. By providing a Model Context Protocol server and agent skills, we aim to let LLM-based agents invoke motion planning directly as a tool.

As robotic systems scale from single manipulators to teams of cooperating robots, multi-robot planning in composite configuration spaces becomes a central challenge. Recent work has shown that sampling-based methods can be extended to these settings by exploiting problem structure~\cite{guo2024efficient}, and future versions of OMPL aim to provide support for multi-robot coordination. At the same time, GPU-parallel approaches to tree construction and collision checking~\cite{huang2025prrtc} offer the potential for massive speedups. Finally, emerging applications in soft and growing robots~\cite{gaochen2025actsim} involve high-dimensional, continuously deformable state spaces that challenge existing representations and will require extending OMPL's state space abstractions to new physical domains.

These directions share a common theme: the core abstractions of sampling-based planning remain the right foundation, but must be extended to meet the demands of increasingly complex robotic systems. We see OMPL~2.0 not as an endpoint, but as a starting point for the next decade of motion planning research.

\ifanon
\section*{Acknowledgements}
Acknowledgements are withheld to preserve anonymity during review.

\else

\section*{Acknowledgements}

The authors gratefully acknowledge the many contributors whose work over the years has shaped the OMPL~2.0. All contributors are acknowledged on the OMPL web page. Contributions can be found in the OMPL GitHub repository. We note, however, that commit counts and lines of code are imperfect indicators of impact. Several significant contributions, such as the initial RRT* implementation, predate the move to GitHub and do not appear in those statistics. The authors would like to thank current and former members of the Kavraki Lab, as well as the broader motion planning community, whose discussions, feedback, and support have been instrumental to this work.
This work was supported in part by NSF OAC-2411219. Theodoros Tyrovouzis and Lydia E.\ Kavraki have been supported in part by ERDC W912HZ2320003. 

\fi

\printbibliography{}

\end{document}